\documentclass[letterpaper, 10 pt, conference]{ieeeconf}  

\IEEEoverridecommandlockouts                              

\usepackage{graphics} 
\usepackage{epsfig} 
\usepackage{amsmath} 
\usepackage{amssymb}  
\usepackage{bm} 

\usepackage[flushmargin,hang]{footmisc} 

\usepackage{flushend} 
\usepackage{comment}	

\usepackage{siunitx} 

\usepackage{pgfplots}
\pgfplotsset{compat=newest}
\usetikzlibrary{plotmarks}
\usetikzlibrary{arrows.meta}
\usepgfplotslibrary{patchplots}
\usepackage{booktabs} 
\usepackage{grffile}
\pgfplotsset{plot coordinates/math parser=false}
\newlength\fwidth
\newlength\fheight

\newcommand{\todo}[1]{{\color{red}\textit{{\textbf{@TODO:}} #1}}}
\usepackage{cite}

\newcommand{\fig}[1]{Fig.~\ref{#1}}

\newcommand{\tab}[1]{Table~\ref{#1}}

\def\epsgaiji#1{\leavevmode\kern-0.025zw\raise-.37zh\hbox{%
  \epsfile{file=#1,width=1.05zw}}\kern-0.025zw}
\newcommand{\MARU}[1]{{\ooalign{\hfil#1\/\hfil\crcr\raise.167ex\hbox{\mathhexbox20D}}}}

\usepackage{array}

\title{\LARGE \bf
Data-Driven Terramechanics Approach Towards a Realistic Real-Time Simulator for Lunar Rovers}

\author{Jakob M. Kern, James M. Hurrell, Shreya Santra, Keisuke Takehana,\\ Kentaro Uno, and Kazuya Yoshida
\thanks{$^{*}$This work was supported by JST Moonshot R\&D Program, Grant Number JPMJMS223B.}
\thanks{All authors are with the Space Robotics Lab (SRL) in the Department of Aerospace Engineering, Graduate School of Engineering, Tohoku University, Sendai 980--8579, Japan.}%
\thanks{
\textit{The corresponding author is Jakob M. Kern.}
    }
\thanks{~~~~(E-mail: \tt{kern.jakob.marian.s4@dc.tohoku.ac.jp})}
    }%

\begin{document}

\maketitle
\thispagestyle{empty}
\pagestyle{empty}


\begin{abstract}

High-fidelity simulators for the lunar surface provide a digital environment for extensive testing of rover operations and mission planning. However, current simulators focus on either visual realism or physical accuracy, which limits their capability to replicate lunar conditions comprehensively.
This work addresses that gap by combining high visual fidelity with realistic terrain interaction for a realistic representation of rovers on the lunar surface. Because direct simulation of wheel-soil interactions is computationally expensive, a data-driven approach was adopted, using regression models for slip and sinkage from data collected in both full-rover and single-wheel experiments and simulations.
The resulting regression-based terramechanics model accurately reproduced steady-state and dynamic slip, as well as sinkage behavior, on flat terrain and slopes up to 20\(^\circ\), with validation against field test results. Additionally, improvements were made to enhance the realism of terrain deformation and wheel trace visualization.
This method supports real-time applications that require physically plausible terrain response alongside high visual fidelity.

\end{abstract} 


\section{Introduction}\label{introduction}

The limited opportunities for testing lunar rovers during real missions highlight the need for realistic lunar surface simulators. A digital environment allows for replicating lunar conditions, ranging from visual features, such as extreme lighting with long shadows, to terrain features. 
However, for planetary rover operations and mission planning, it is essential that the simulated lunar environment not only appears visually realistic but also accurately models the physical interaction between the rover and the terrain. Especially, the slip ratio, a measure of wheel traction, and wheel sinkage into the terrain are critical features that occur on granular soil, such as the lunar regolith. While the wheel-soil interaction can be analyzed in detail using contact dynamics and particle simulations, this approach is computationally intensive and not suitable for real-time applications. 

\begin{figure}[t]
    \centering
    \includegraphics[width=\linewidth]{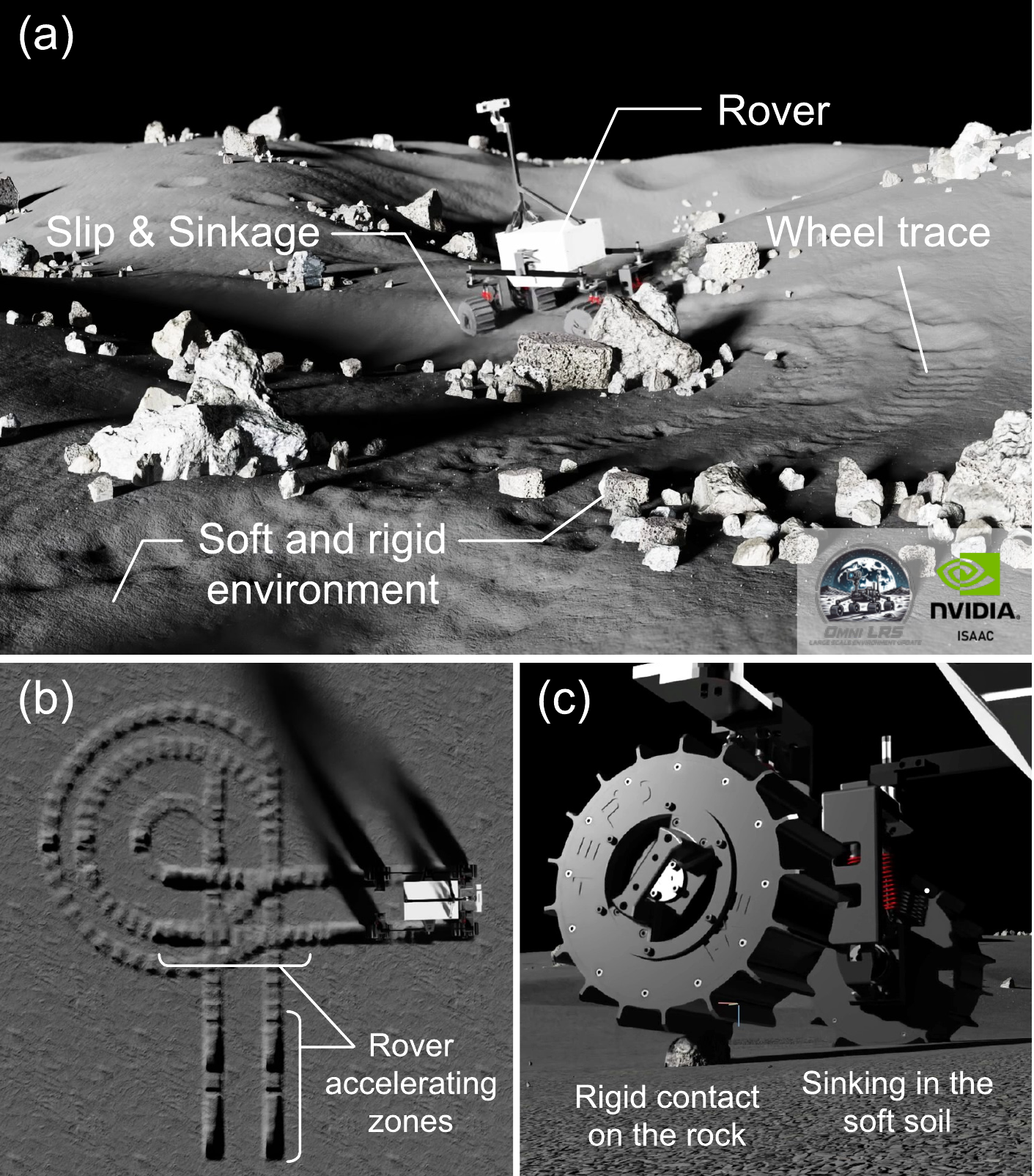}
    \caption{(a) Data-driven terramechanics model for recreating realistic wheel slip, sinkage and wheel traces inside the real-time lunar surface simulator OmniLRS. (b) A rover makes physically reasonable wheel traces in our simulator. (c) Rigid and soft contacts are treated respectively.}
    \label{fig:simulator}
\end{figure}

To address this limitation, this study employs a data-driven approach to enhance the physical fidelity of wheel-terrain interactions for real-time lunar surface simulations, as shown in \fig{fig:simulator}. Regression models for slip ratio and sinkage were derived from experiments conducted with a high-speed rover \cite{EX1} at the Space Exploration Field in the Japan Aerospace Exploration Agency (JAXA) Sagamihara Campus \cite{Tansa_X}. These models were integrated into a rigid-body simulation framework to replicate realistic wheel slip and sinkage in a digital environment while maintaining low computational costs. The main contributions of this work are:
\begin{itemize}
    \item Wheel slip implementation, based on wheel velocity and slope angle
    \item Wheel sinkage implementation, based on slip ratio and wheel load
    \item Terrain deformation, based on slip ratio, to improve the realism of rover wheel traces
\end{itemize}

This work is integrated into the open-source OmniLRS simulator \cite{JUN}, which is based on NVIDIA IsaacSim \cite{IsaacSim}, and extends its capabilities for planetary rover testing with improved physics fidelity.


\section{Related Works}\label{relatedwork}

High-fidelity simulation of lunar rover mobility requires both visually convincing environments and physically accurate wheel-soil interactions. However, existing simulation tools tend to prioritize one at the expense of the other. Real-time lunar surface simulators typically emphasize high visual fidelity and integration with robotic control frameworks, but they often simplify terrain interactions. On the other hand, high-fidelity terramechanics simulators model granular soil dynamics in detail but are too computationally intensive for real-time use. In this section, we review key works from both areas and motivate our approach, which bridges the gap between them.

\subsection{Real-time Lunar Surface Simulators}

Several high-fidelity lunar surface simulators for real-time applications are actively being developed, with a primary focus on visual realism. The lunar environments use rigid terrain models and simple physics models, limiting their utility for realistic rover mobility simulation.

One example is the \textbf{Digital Lunar Exploration Sites (DLES)} \cite{DLES_1,DLES_2} developed by NASA's Johnson Space Center, which provides a detailed model of the lunar South Pole, derived from the Lunar Orbiter Laser Altimeter (LOLA) \cite{LOLA-1}.
The authors describe the processing of digital elevation maps and the efficient handling of large-scale environments, including craters and boulders. While DLES was designed for use across multiple rendering engines, the \textbf{DLES Unreal Simulation Tool (DUST)} \cite{DUST} integrates the DLES terrain database into Unreal Engine 5. It features high-resolution terrain, realistic lighting, and tools for planning rover traverses. However, it lacks terrain deformation modeling and complex wheel-terrain dynamics. Wheel traces are visually rendered, but not physically simulated.

The \textbf{RSIM Lunar Surface Simulator} \cite{RSIM} (formerly known as the VIPER Lunar Surface Simulator \cite{VIPER}), developed by NASA's Ames Research Center and built on top of Gazebo, features many crucial aspects of lunar exploration. The simulator includes realistic lighting, camera lens flare and synthetically enhanced large-scale terrain. It consists of a slope-dependent wheel slip model, derived at the NASA Glenn SLOPE facility, which uses a 'slip map' to simulate loss of traction. However, this model does not consider velocity-dependent slip behavior, and wheel sinkage due to slip is not visually or physically represented. Wheel traces are rendered using bump mapping, which creates the illusion of deformation by altering surface normals; however, it does not modify the actual terrain mesh. The trace rendering does not consider wheel geometry or slip.

\textbf{OmniLRS} is an open-source lunar robotic simulator based on NVIDIA Omniverse IsaacSim and was developed by the University of Luxembourg and the Space Robotics Lab at Tohoku University \cite{JUN}. It features three environments: (1) the Lunalab, a replica of the lunar testbed located at the University of Luxembourg, (2) the Lunaryard, a procedurally generated lunar terrain, and (3) a large-scale lunar environment based on lunar digital elevation maps. The simulator has demonstrated successful sim-to-real transfer in vision-based rock segmentation tasks using the Lunalab environment. Wheel traces are visually represented and scaled through a contact-force-based regression model, which can be tuned to suit a grouser wheel if the mesh geometry of the wheel is available~\cite{OmniLRS_deformation}. However, the simulator lacks full terramechanics considerations, such as a slip plugin or wheel sinkage visualization.

\subsection{Terramechanics Simulators}

Terramechanics simulators prioritize the accurate computation of interactions between vehicles and terrain over the realistic visual representation of synthetic environments. In addition, typically, rover perception and control are either simplified or lacking. 

For high-fidelity modeling, the discrete element method (DEM), originating with the works of Cundall and Strack \cite{DEM}, can be used to model lunar soil as a collection of particles that interact with one another and a wheel of the vehicle traversing the terrain. Contact forces and torques are calculated using computationally intensive methods with high spatial and temporal resolution. DEM has been widely used in terramechanics to study slip, wheel sinkage and soil deformation. Coupling DEM with multi-body dynamics enables detailed analyses of planetary rover mobility on granular material, ranging from complex single-wheel motion and suspension systems to full rover studies \cite{DEM_multibody}. However, the method is not suitable for real-time applications and lacks support for robotic control and sensors.

To address this limitation, frameworks such as Project Chrono \cite{Chrono_1} offer multiple deformable terrain models. While supporting DEM, Chrono has a continuous representation model (CRM) and a soil contact model (SCM). SCM computes contact forces between surfaces and the ground, allowing for terrain mesh deformation and contact response in near real-time. However, simulations are restricted to small environments and lack integration with robotics stacks for perception and control. 
\section{Methodology}\label{methodology}
\begin{figure}[t]
    \centering
    \includegraphics[width=1.0\linewidth]{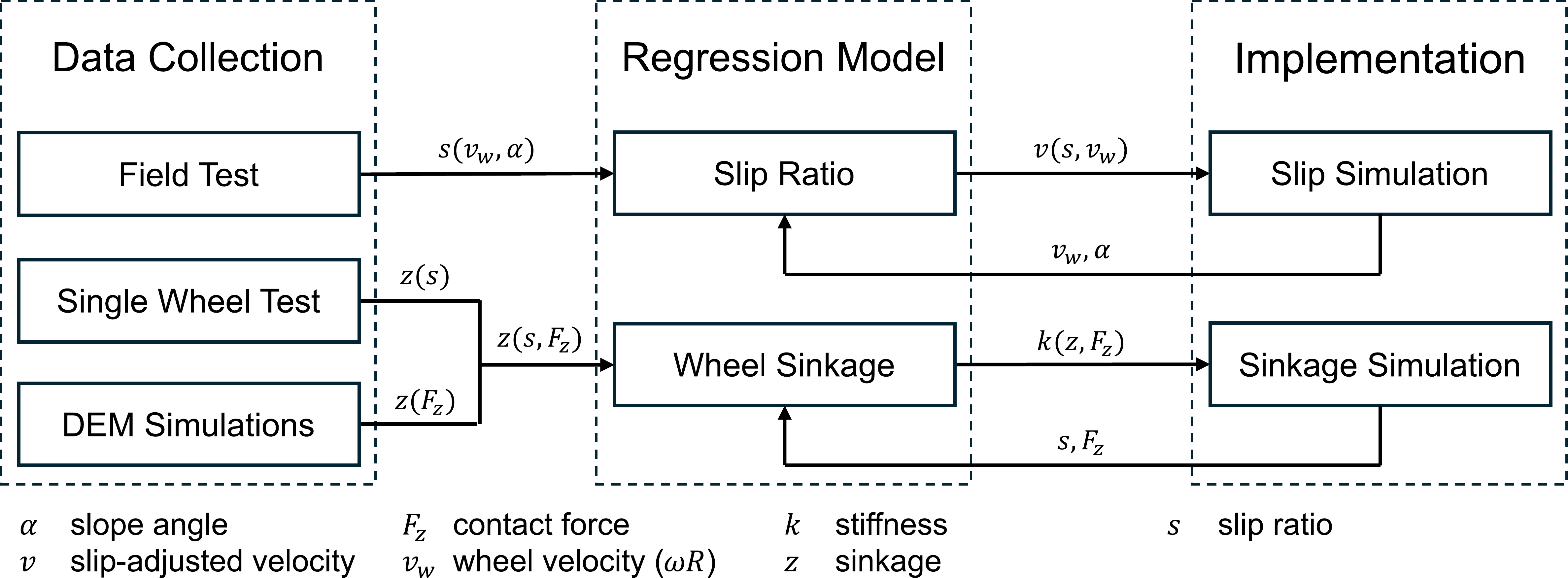}
    \caption{Architecture of the proposed method.}
    \label{fig:methodology}
\end{figure}
This work aims to bridge high-fidelity terramechanics simulations and real-time photorealistic robotics environments using a data-driven approach, as illustrated in \fig{fig:methodology}. Regression models for wheel slip and sinkage, derived from field tests, single-wheel experiments, and DEM simulations, are integrated into the lunar surface simulator OmniLRS \cite{JUN}. Slip ratio \(s\) is simulated by commanding the rover to follow a slip-adjusted velocity \(v\), while sinkage \(z\) is modeled via compliant contacts (spring-damper system), where stiffness \(k\) determines the sinkage depth. In addition, the contributions include improved terrain deformation for wheel trace rendering that not only scales with the contact force \(F_z\), as implemented in \cite{OmniLRS_deformation}, but additionally with the slip ratio \(s\). This paper focuses on the implementation method rather than on data collection or regression modeling. 

\subsection{Experiment Procedure}\label{experiment_procedure}

\begin{figure}
    \centering
    \includegraphics[width=1.0\linewidth]{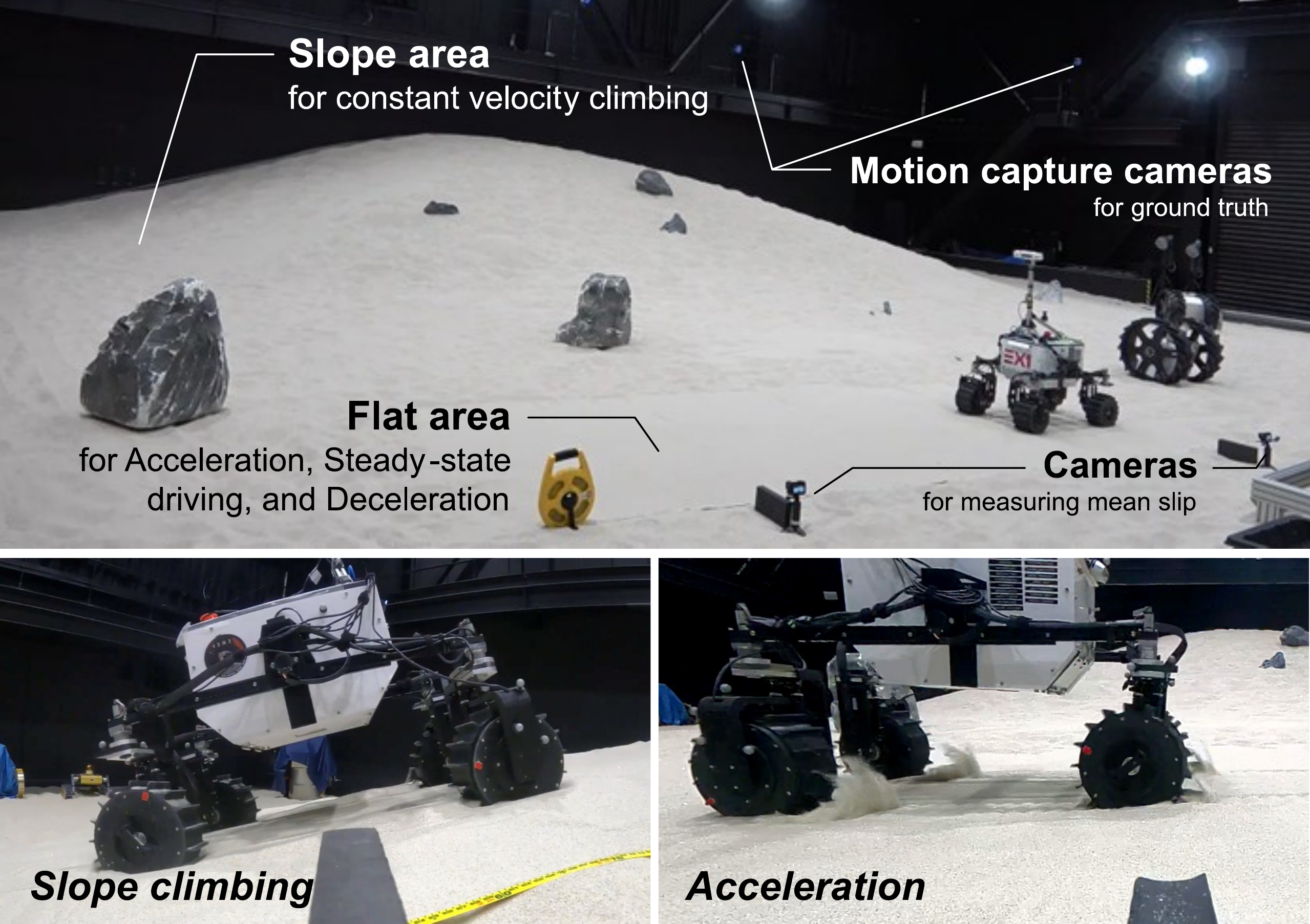}
    \caption{Experimental setup of the data acquisition field test (top) and two representative snaps of slope climbing (bottom-left) and acceleration (bottom-right).}
    \label{fig:field_experiments}
\end{figure}
\begin{figure}
    \centering
    \includegraphics[width=1.0\linewidth]{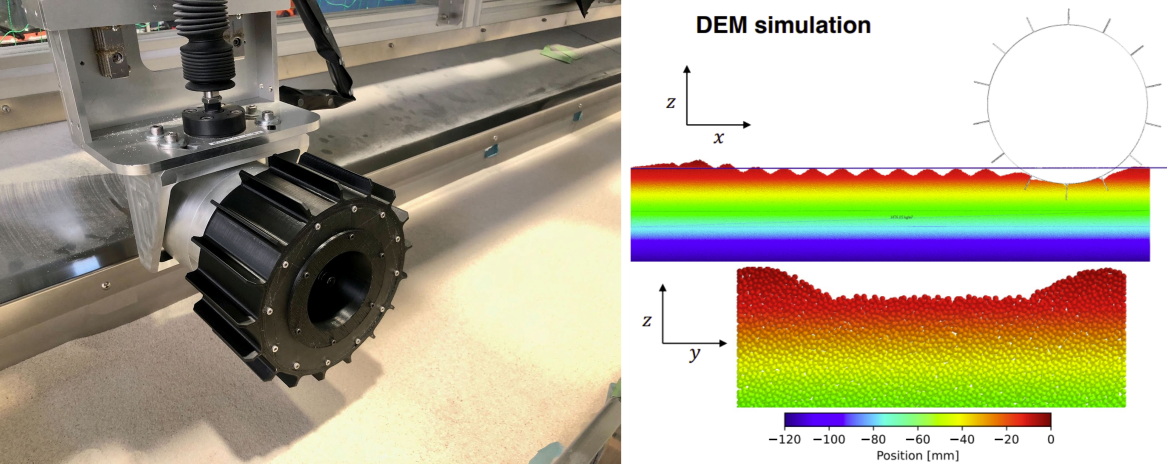}
    \caption{Single wheel testbed (left) \cite{EX1_grouser} and DEM simulation (right) \cite{Hurrell_2}.}
    \label{fig:lab_experiments}
\end{figure}
\begin{table*}[bthp]
\vspace{1.8mm}
\centering
\caption{Overview of sensors, data, sampling frequency, and number of data points collected for slip ratio and sinkage estimation}
\label{tab:data_collection}
\begin{tabular}{ccrrr}
    \hline
    \textbf{Sensors} & \textbf{Recorded Data} & \textbf{Computed Data} & \textbf{Frequency} & \textbf{Data Points}\\
    \hline
    Motion Capture System & Position $[x,y,z]$ and Orientation $[q_x,q_y,q_z,q_w]$
 & Velocity $ v $ & $180 \text{Hz}$ & $\leq1000$\\
    Cameras     & Images & Avg. Velocity $ v $ & \(60 \text{Hz}\) & - \\
    Motor Encoders & Angular velocity $[\omega]$
 & - & \(5 \text{Hz}\) & $\leq35$\\
    \hline
\end{tabular}
\end{table*}
To measure the wheel slip, field tests were conducted at the JAXA Space Exploration facility \cite{Tansa_X}, using EX1, a four-wheeled high-speed rover of dimensions ($0.82,0.52,0.67\ \text{m}$) and weighing \SI{21.63}{kg}, equipped with a passive spring-damper suspension developed at the Space Robotics Lab \cite{EX1}. The wheel design is optimized based on the grouser spacing equation in \cite{grouser}. The data collection test was designed as shown in \fig{fig:field_experiments}. The experimental site 
features a $\SI{20}{m} \times \SI{20}{m}$ area filled with of Tohoku Silica No.5 \cite{Silica} sand of about \SI{0.3}{m} in depth. This dry loose silica sand is commonly used in terramechanics research as a low-fidelity lunar regolith simulant. Its sparse grain distribution presents even more challenging conditions for wheel sinkage and slippage \cite{EX1_grouser}. 

Tests were performed on flat terrain across wheel velocities (\(v_w = \omega R\)) of \(0.23\) to \(1.17\ \text{m/s}\), and on slopes ranging from \(0^\circ\) to \(18^\circ\) at speeds up to \(0.47\ \text{m/s}\). Additional trials included rapid acceleration and deceleration to analyze transient slip (Fig.~\ref{fig:field_experiments}). Wheel speed \(v_w\) was measured via encoders, and translational velocity \(v\) was tracked using a 16-camera OptiTrack motion capture system (\(180\ \text{Hz}\), \(\pm\ 0.15\ \text{mm}\) accuracy). To independently estimate average slip, external cameras recorded wheel rotation and traversal time over a defined \SI{2}{m} section (also see \tab{tab:data_collection}). 

Slip ratio \(s\) was computed as shown in Equation~\ref{equ:slipratio}, where \(\omega\) is the angular velocity and \(R\) the effective wheel radius, including grouser height. In many studies of grousered wheel motion, only the wheel radius is used, ignoring grouser height for slip ratio calculations. However, it was found that if the grouser length were not included for slip calculations, then the slip would be calculated as negative or skidding during the driving motion. Including the grouser length as the effective wheel radius yields positive slip for all acceleration and steady-state motion. The slip or skid is then negative, as expected, during deceleration.
\begin{equation}
    s = 
    \begin{cases}
        1- \frac{v}{\omega R}, \quad \text{if } \omega R \geq v \\
        \frac{\omega R}{v} - 1, \quad \text{if } \omega R < v \\
    \end{cases}
\label{equ:slipratio}
\end{equation}
During acceleration and deceleration tests, the motion capture system also tracked vertical wheel displacement to estimate sinkage relative to the initial resting position.
To complement the field data, prior results from single-wheel testbed experiments \cite{EX1_grouser} and DEM simulations \cite{Hurrell_1,Hurrell_2} were referenced (Fig.~\ref{fig:lab_experiments}), as DEM enables detailed analysis of wheel–soil interaction where physical testing is limited by controllability and measurement precision.
The testbed provided slip–sinkage relationships, while DEM simulations at a fixed slip ratio of \(20\%\) were used to evaluate the influence of varied vertical loads. 

\subsection{Regression Models}\label{regression_model}

Based on the experimental observations, regression models for wheel slip and sinkage were derived to characterize terrain interactions. The slip ratio \(s\) on flat terrain was found to increase linearly with wheel speed \(v_w\), as expressed in Equation~\ref{equ:slip_vel}. The effect of slope angle \(\alpha\) on slip ratio was captured by fitting a second-order polynomial with coefficients parameterized by \(v_w\), resulting in the combined slip model shown in Equation~\ref{equ:slip_angle}. The polynomial fits were based on the averaged values from 16 independent experimental runs.
\begin{equation}
    s(v_w) = 0.0265\ v_w + 0.0256
    \label{equ:slip_vel}
\end{equation}
\begin{equation}
    s(v_w.\alpha) = (0.00522\ v_w + 0.00105)\ \alpha^2 + s(v_w)
    \label{equ:slip_angle}
\end{equation}

The sinkage model was developed by combining results from the single-wheel testbed and discrete element method (DEM) simulations. Sinkage \(z\) was modeled as a linear function of slip ratio \(s\) and the deviation of vertical load \(F_z\) from a nominal reference load \(F_{ref}\), which corresponds to the static load on a single wheel under uniform force distribution in lunar gravity. Static sinkage was measured by the motion capture system during the field tests. The sinkage regression model, where \(z\) is expressed in millimeters (mm), is given by:
\begin{equation}
    z(s, F_z) = -33.56\ s - 0.9291 (F_z - F_{ref}) - 3.11
\end{equation}

\subsection{Slip Implementation} \label{slip_implementation}

In both real-world operation and high-fidelity terramechanics simulations, slip arises from complex interactions between the wheels and deformable terrain. In contrast, the present approach determines wheel slip using a regression model. At each simulation time step, the slip ratio \(s\) is computed by evaluating the regression model with the current wheel velocity \(v_w\) and local slope angle \(\alpha\) as inputs. This slip ratio is then used to determine the slip-adjusted translational velocity \(v\), defined as:
\begin{equation}
    v = \left( 1-s(v_w, \alpha)\right)\ v_w
\end{equation}

Since the simulation is based on rigid-body dynamics with Coulomb friction, it cannot reproduce steady-state slip, i.e., slip occurring during motion with a constant velocity. To replicate the kinematic effect of slip, the wheels are directly actuated with the slip-adjusted velocity \(v\). This approach enables the realistic simulation of slip behavior while maintaining a rigid-body framework.

\begin{figure}
    \centering
    \includegraphics[width=1.0\linewidth]{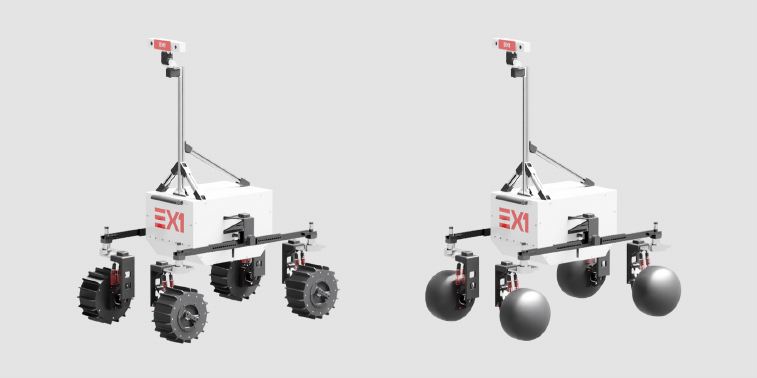}
    \caption{Rover illustrated with visual wheel geometry (left) and physics wheel geometry (right).}
    \label{fig:slip_impl_geometries}
\end{figure}

To reproduce the visual appearance of slipping wheels, each wheel is separated into a purely visual body for rendering and a purely physical body for physics interactions. Both representations share the same location but are connected to the chassis via separate joints, allowing independent actuation. The physical wheel is driven at the slip-adjusted velocity \(v\), while the visual wheel rotates at the commanded wheel velocity \(v_w\). A visual impression of wheel slip is then created, without actual slip occurring between the physics geometry and the terrain.
While the visual geometry utilizes a detailed mesh for realistic representation of the grouser wheel, the physics geometry is based on a simple capsule (\fig{fig:slip_impl_geometries}) to achieve stable contact interactions.

In addition to steady-state slip behavior, the rover's transient dynamics during acceleration and deceleration are essential for consistency with experimental data. Accordingly, the simulation Coulomb friction model was calibrated using velocity profiles obtained from field tests. These tests showed that the rate of deceleration remained consistent from \SI{0.2}{m/s} to \SI{1.17}{m/s} for commanded velocities during rapid stops. By setting the static and dynamic friction coefficients to 1.0 and 0.8, respectively, the simulated deceleration closely matched the observed behavior.

\begin{figure}
    \centering
    \includegraphics[width=1.0\linewidth]{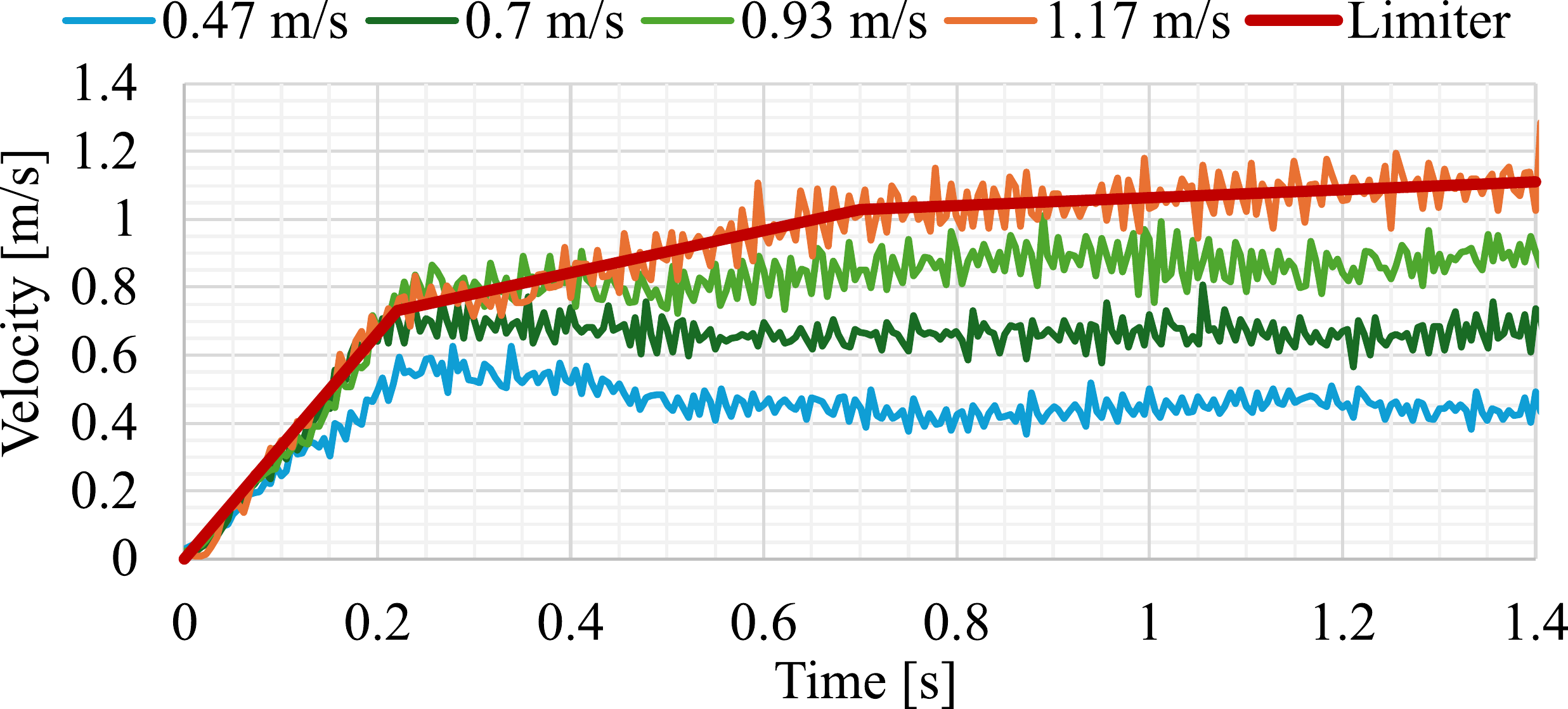}
    \caption{Measured translational velocity during acceleration for four different commanded wheel velocities. The red curve represents the limiter function, which defines the maximum possible acceleration.}
    \label{fig:slip_implementation_acc}
\end{figure}

However, the high friction parameters lead to unrealistically rapid acceleration. To address this, we introduced a limiter function that constrains the maximum allowable change in translational velocity between time steps. This constraint is based on field tests, where initial acceleration was largely independent of the commanded wheel velocity, indicating that it is limited by grouser geometry and soil characteristics. The field test data and the limiter are shown in \fig{fig:slip_implementation_acc}.
Notably, acceleration decreased beyond a certain velocity threshold, a trend that can be approximated using a piecewise linear function, defined as follows:
\begin{equation}
    \Delta v \leq 
    \begin{cases}
        3.476, & \text{if } v  \leq 0.75 \\
        0.612, & \text{if } 0.75 < v  \leq 1.02 \\
        0.114, & \text{if } v  > 1.02
    \end{cases}
\end{equation}

\subsection{Sinkage Implementation} \label{sinkage_implementation}

To model wheel sinkage on deformable terrain using only rigid bodies, we use compliant contact modeling. Rather than treating contacts as perfectly rigid, a spring-damper system simulates the vertical contact force between colliding bodies, the terrain and the wheels. The system is expressed as:
\begin{equation}
    \ddot{z}(t) = -g - \frac{k}{m}z(t) - \frac{c}{m} \dot{z}(t) \quad \text{for} \quad z(t)<0 \\
\end{equation}

Here, \(z(t)\) is the penetration depth, \(\dot{z}(t)\) the relative normal velocity, and \(\ddot{z}(t)\) the relative acceleration of the two bodies. The parameters \(k\) and \(c\) are the stiffness and damping coefficients, respectively, \(m\) is the effective mass at the contact, and \(g\) is gravitational acceleration. The vertical reference contact force is \(F_z = mg\).

In quasi-static conditions (i.e., \(\dot{z}(t) = \ddot{z}(t) = 0\)), the penetration depth can be obtained by:
\begin{equation}
    z = -\frac{F_z}{k}
\end{equation}

Thus, for a vertical force \(F_z\), the penetration depth \(z\) can be controlled by adjusting the stiffness \(k\). Since our regression model provides the expected sinkage \(z(s, F_z)\) as a function of slip ratio \(s\) and vertical load \(F_z\), we invert the relation to determine the appropriate stiffness. 

For grouser wheels, sinkage is typically measured relative to the base radius \(r\), not considering sinkage of the grousers. When simulating grouser wheels using a physics geometry with radius \(R = r + h\), the grouser height \(h\) must be explicitly considered in the equation to achieve the correct sinkage.



Collisions between the wheels and the terrain may be resolved at a single or multiple contact points, with the total load distributed among these points. To achieve the correct penetration depth \(z\), the stiffness \(k\) must be scaled by the number of contact points \(N\), as shown in Equation~\ref{equ:k}. It is advisable to use simple physics geometries for rover wheels, such as spheres or capsules, that produce steady contact forces through a consistent number of contact points. 
\begin{equation}
    k = -\frac{1}{N}\frac{F_{z}}{z(s,F_z)-h}
    \label{equ:k}
\end{equation}

For numerical stability, minimum thresholds are enforced for both the contact force \(F_z\) and the denominator in the stiffness calculation. This prevents divisions by zero and avoids near-zero stiffness values that could otherwise lead to unstable behavior when \(F_z\) is small.

The stiffness parameter \(k\) is updated at each timestep based on the current contact force and slip ratio, which can lead to unrealistic behavior during deceleration. As the rover skids, high slip causes deep sinkage. Once it stops and slip drops to zero, the recalculated sinkage becomes shallower, causing a sudden increase in \(k\) and making the rover rise visibly. To avoid this artifact, sinkage is constrained when the translational velocity \(v\) falls below \(v_{min}=0.1\ \text{m/s}\):
\begin{equation}
    z^{(t)} \leftarrow \min(z^{(t-1)}, z^{(t)}), \quad \text{if } v^{(t)} \leq v_{\min}
\end{equation}

Complementing the stiffness parameter, the damping constant \(d\) affects the dynamic response of the contact. In this work, the damping constant is set to critical damping (Equation~\ref{c}), which ensures stable behavior:
\begin{equation}
    c = 2 \sqrt{km}
    \label{c}
\end{equation}

\subsection{Terrain Deformation}\label{terrain_defomation}

The terrain implementation by Kamohara et al. \cite{OmniLRS_deformation} is extended by incorporating slip, in addition to contact forces, as a second parameter influencing the shape and depth of wheel traces. Higher slip values result in deeper terrain deformation, and the amplitude of the sinusoidal trace pattern left behind by grouser wheels is reduced under high slip conditions, which is consistent with field test observations. 
In addition to the geometric adjustment of the wheel, the deformation mechanism was separated into two components: permanent depth deformation and transient trace patterns. This separation enhances the visual fidelity of overlapping wheel traces, particularly when the rover traverses the same area multiple times.

In the current version, depth deformation is only updated if the newly computed depth \(d\) is greater (i.e., deeper) than the existing value at a given terrain cell \(d_{x,y}\). In contrast, the trace pattern \(w\) is regenerated and applied on every pass, allowing for realistic wheel trace visuals. The combination of depth \(d\) and trace \(w\) results in realistic and dynamic terrain modifications in the digital elevation map. The update logic is defined as:
\begin{align*}
    \Delta d &\leftarrow \min(0, d - d_{x,y}) \\
    \Delta w &\leftarrow w - w_{x,y} \\
    DEM_{x,y} &\leftarrow DEM_{x,y} + \Delta d + \Delta w \\
    d_{x,y} &\leftarrow d_{x,y} + \Delta d \\
    w_{x,y} &\leftarrow w
\end{align*}

\subsection{Simulation Setup}
The implementation was developed using the robotics simulator IsaacSim (version 2023.1.1) in combination with the OmniLRS environment. The slip and sinkage models were implemented as Omnigraphs, primarily utilizing custom script nodes. The wheel trace functionality was directly integrated into the OmniLRS base code. All simulations were performed in the Lunaryard environment with default physics settings \cite{JUN} and a simulation timestep of \(30\ \text{Hz}\).

\section{Results and Analysis}\label{results}
\begin{figure}[thb]
    \centering
    \includegraphics[width=\linewidth]{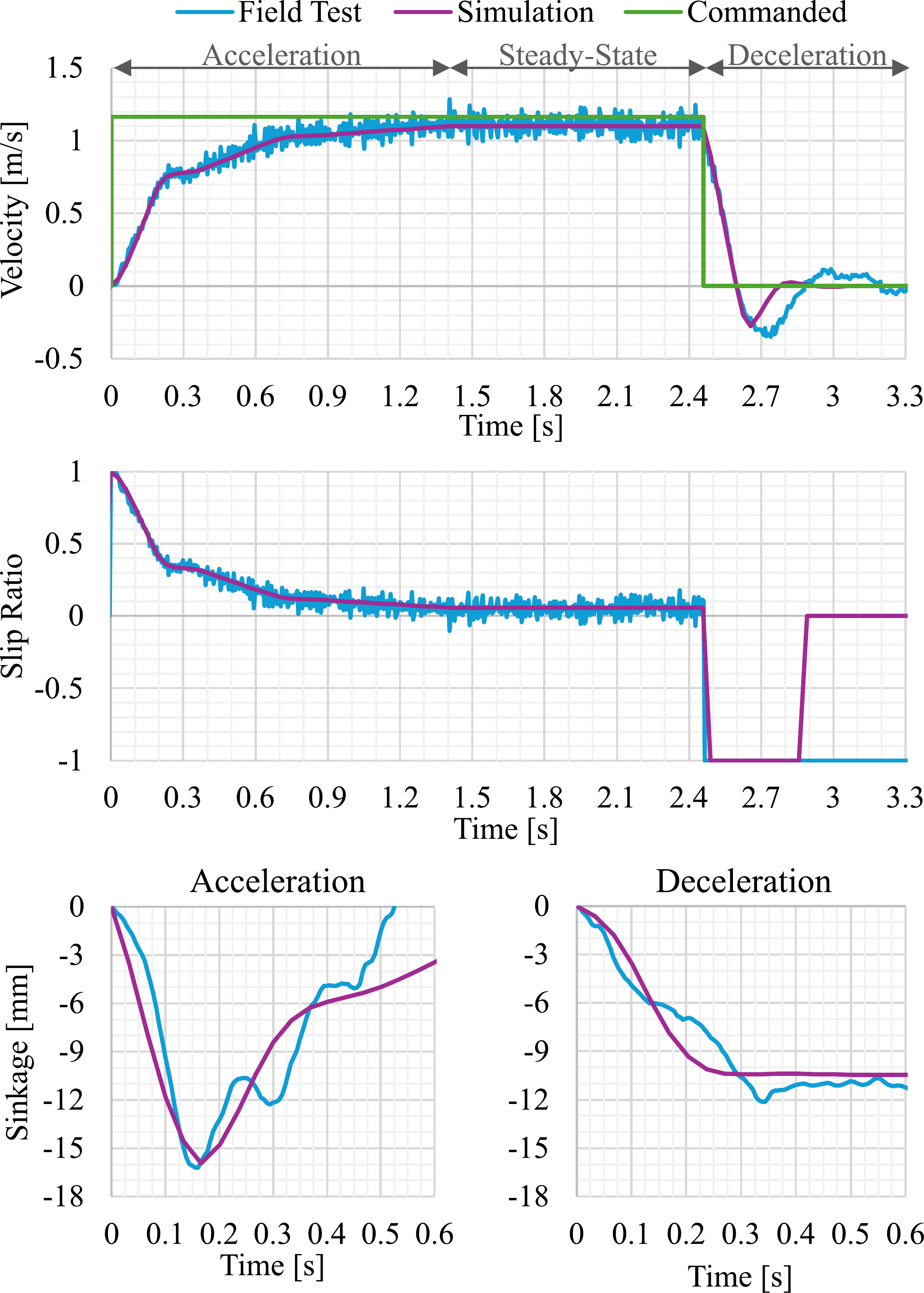}
    \caption{Comparison between results from field test and simulation for the same experiment at a commanded wheel velocity of 1.17 m/s, including sudden acceleration and deceleration phases.}
    \label{fig:results}
\end{figure}

\begin{figure}[thb]
    \centering
    \includegraphics[width=\linewidth]{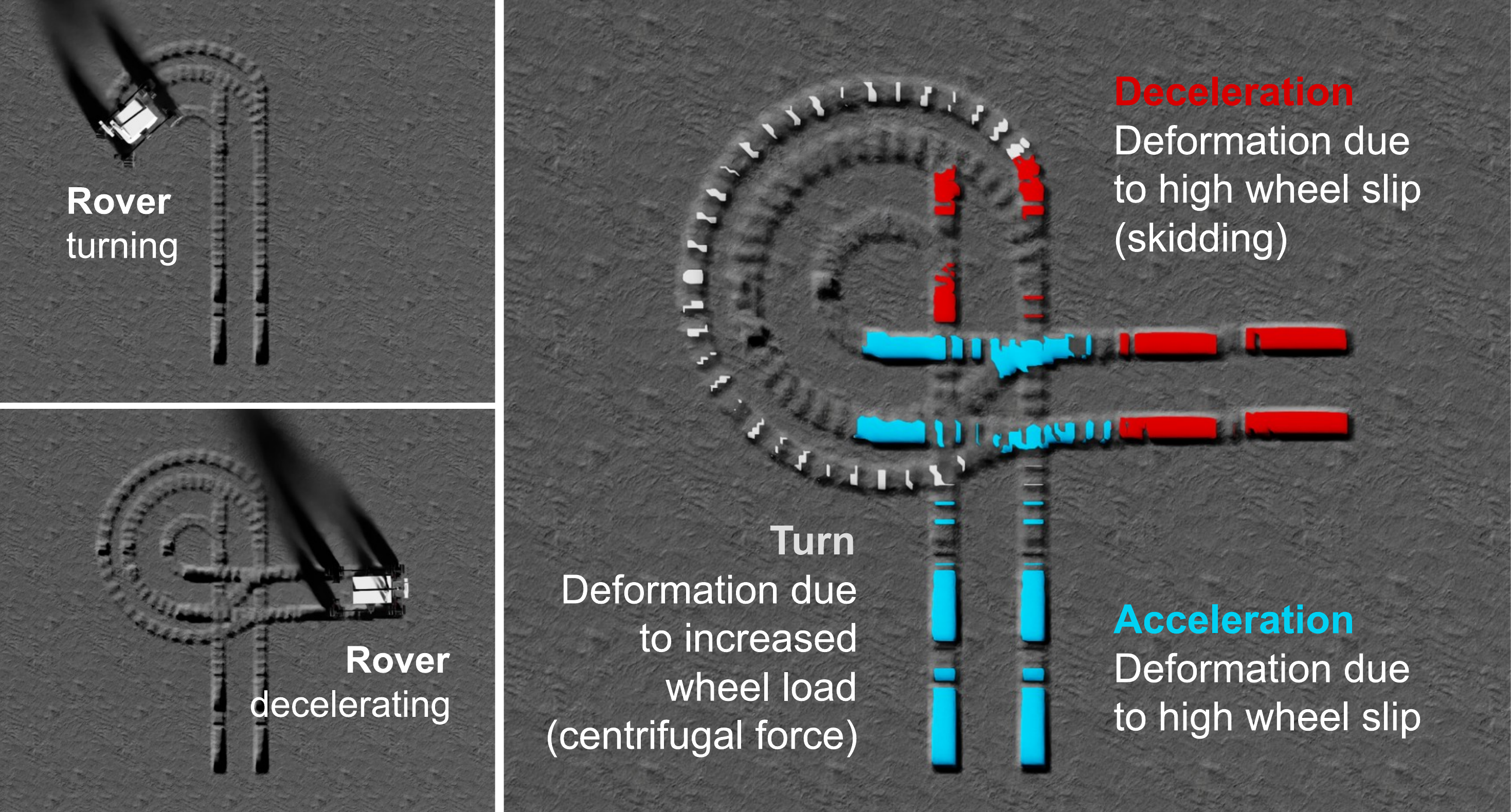}
    \caption{Visible terrain deformation after rover traversal, with highlighted areas indicating locations where deformation depth is greater than \(12\ \text{mm}\).}
    \label{fig:result_deformation}
\end{figure}

\begin{figure*}[thb]
    \centering
    \includegraphics[width=\linewidth]{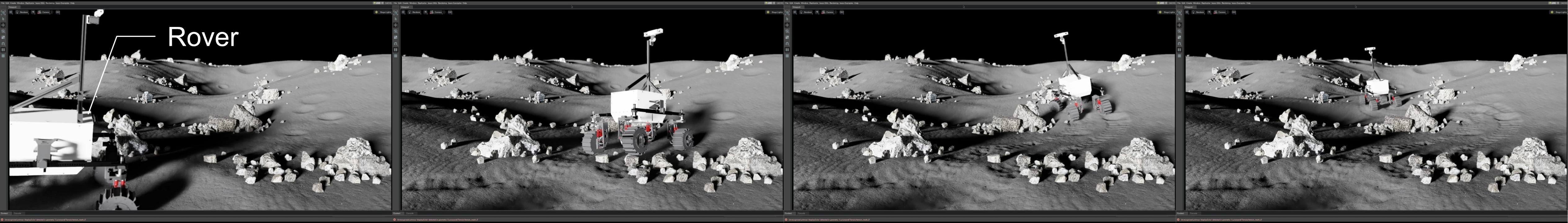}
    \caption{Snapshots of our completed simulation platform, showing the four-wheeled EX1 rover traversing the digital lunar environment. Wheel slip and sinkage change dynamically with the terrain (slope angle) and the rover’s motion (wheel velocity, contact forces) according to the regression models. Wheel traces are visualized through real-time deformation of the visual terrain mesh. When traversing rocks, the very high compliant contact stiffness assigned to them effectively prevents any wheel sinkage. 
    }
    \label{fig:sim_overview_result}
\end{figure*}
The implementation was evaluated by comparing simulation results for slip ratio and wheel sinkage against the outputs of the regression models, providing a measure of how accurately the simulation reproduces expected behavior. 

On flat terrain, the slip ratio error remained below \(0.02\%\) across all wheel speeds up to \(1.17\ \text{m/s}\) on flat terrain. For inclined terrain with slope angles up to \(20^\circ\), the mean absolute error averaged to \(0.22\%\). The average error in sinkage depth was \(0.25\ \text{mm}\) across the full range of slip ratios and contact force variations within \(\pm 5\ \text{N}\) of the reference force of \(8.72\ \text{N}\) under lunar gravity.




Additionally, \fig{fig:results} directly compares the simulation to the field test results on flat terrain for a commanded wheel velocity of \(v_w = 1.17\ \text{m/s}\), showing close agreement during steady-state motion, acceleration, and deceleration. The top graph shows \(v_w\) and the measured translational velocities \(v\). The middle graph compares slip ratios, and the bottom plots display sinkage depth relative to the start of acceleration (left) and deceleration (right).

\fig{fig:result_deformation} shows a top-down view of wheel traces generated using the improved terrain deformation model. Highlighted areas indicate deformation depths exceeding \(12\ \text{mm}\): blue and red mark regions of high slip during acceleration and deceleration, while white corresponds to turning-induced deformation from increased load on the outer front wheel.

Finally, our simulator operates as shown in \fig{fig:sim_overview_result}, achieving realistic wheel slippage and sinkage, as well as physical deformation of the terrain mesh, i.e., wheel traces on the lunar surface.

The proposed method accurately reproduced slip and sinkage behavior, confirming the effectiveness of using a regression-based terramechanics model. Although tuned for a specific rover–terrain combination, it is compatible with other configurations by substituting the regression model with one derived from new experimental data.

While slip was considered at various velocities, on upward and downward slopes, and during acceleration and deceleration on flat terrain, side slip and wheel-dependent slip during turns were not explicitly addressed. Furthermore, the current regression models are based on dry silica sand, which exhibits different and more uniform mechanical properties than lunar regolith. Developing regression models using actual lunar regolith simulants or in-situ lunar data, potentially with online adaptation methods, represents an important direction for future work to improve applicability to real missions. 



\section{Conclusions}\label{conclusion}

This work presented a data-driven approach to replicating realistic rover traversal on lunar terrain in real-time simulations. By integrating a regression-based terramechanics model, the implementation accurately captured both steady-state and dynamic slip and sinkage behavior, showing strong agreement with field test results. It also improved the realism of terrain deformation and wheel track visualization.

The method supports real-time applications where physically realistic terrain interaction responses are important, such as mobility analysis, rover testing, mission planning, or digital twin environments. Future work should focus on adapting to legged robot surface interaction to enhance simulation fidelity across diverse platforms used in space robotics.



\section*{Acknowledgement}
The authors would like to thank Simon Giel, Momoko Shimizu, Antoine Jonquieres, Takeaki Komine, Yoshimasa Muneishi and Nette Levijoki for their invaluable support in field testing and data collection.

\bibliographystyle{IEEEtran}
\bibliography{reference.bib}

\end{document}